\documentclass[final]{cvpr}

\usepackage{times}
\usepackage{epsfig}
\usepackage{graphicx}
\usepackage{amsmath}
\usepackage{amssymb}

\usepackage{float}
\usepackage{bbm}
\usepackage{booktabs}
\usepackage{pifont}
\usepackage{multirow}
\usepackage{threeparttable}
\usepackage{caption}
\usepackage{subcaption}
\usepackage{times} 
\usepackage{enumerate}

\usepackage{algorithm}
\usepackage{comment}
\usepackage{setspace}
\usepackage{algpseudocode}
\usepackage{amsmath}

\usepackage[pagebackref=true,breaklinks=true,colorlinks,bookmarks=false]{hyperref}

\def\cvprPaperID{} 
\def\confYear{CVPR 2021}

\begin{document}

\title{Momentum$^2$ Teacher: Momentum Teacher with Momentum Statistics\\ for Self-Supervised Learning}

\author{Zeming Li, ~~~Songtao Liu, ~~~Jian Sun\\
MEGVII Technology\\
{\tt\small \{lizeming, liusongtao, sunjian\}@megvii.com}
}

\maketitle
\begin{abstract}

In this paper, we present a novel approach, Momentum$^2$ Teacher, for student-teacher based self-supervised learning. The approach performs momentum update on both network weights and batch normalization (BN) statistics. The teacher's weight is a momentum update of the student, and the teacher's BN statistics is a momentum update of those in history. The Momentum$^2$ Teacher is simple and efficient. It can achieve the state of the art results (74.5\%) under ImageNet linear evaluation protocol using small-batch size(\eg, 128), without requiring large-batch training on special hardware like TPU or inefficient across GPU operation (\eg, shuffling BN, synced BN). Our implementation and pre-trained models will be given on GitHub\footnote{https://github.com/zengarden/momentum2-teacher}.

\end{abstract}

\section{Introduction} \label{sec:intro}

Student-teacher framework is one of the key ingredients in current state-of-the-art self-supervised visual representation learning, (\emph{e.g.}, MoCo~\cite{mocov1,mocov2}, BYOL~\cite{BYOL}). The framework learns two networks with the same architecture, where the student network is trained with gradient back-propagation and the teacher network is a momentum version of the student network. 

\begin{figure*}[!tp]
\centering
\includegraphics[width=1\linewidth]{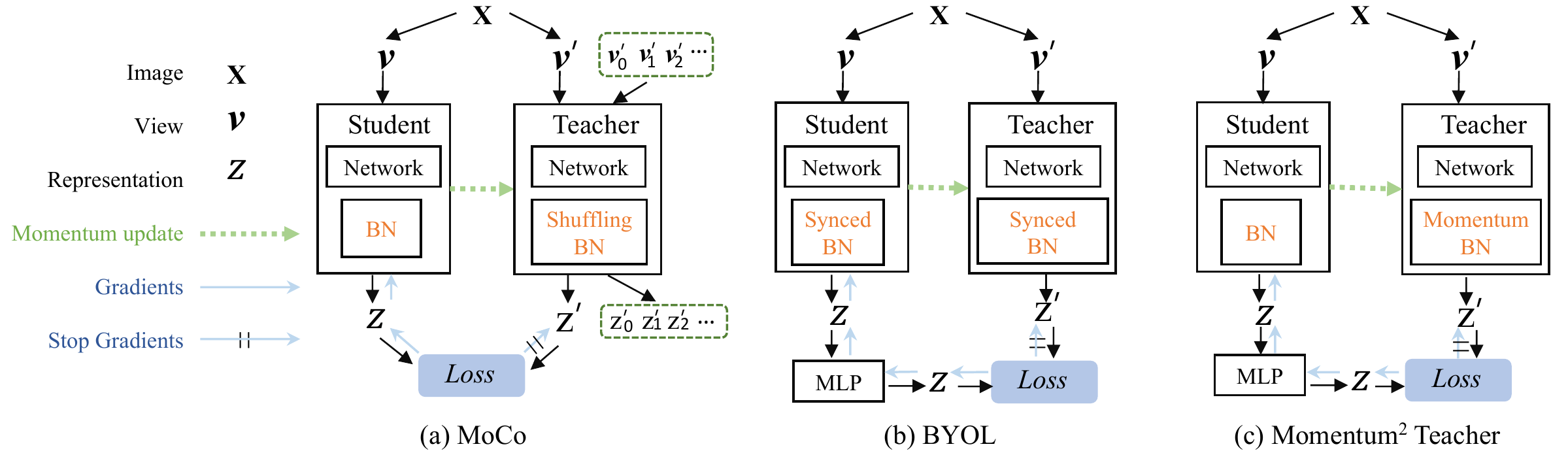}
\caption{Pipeline of student-teacher based self-supervised approaches. The positive pairs~($v, v'$), the different augment views of the same image, are fed into student and teacher networks respectively to learn the representations~($z, z'$). In MoCo~\cite{mocov1}, the student is trained with (small-batch) BN and the teacher uses shuffling BN. ($v'_0, v'_1, v'_2, ...$) are negative pairs from other images. In BYOL~\cite{BYOL}, both student and teacher are trained with large batch and BN statistics with synchronized batch-normalization~(Synced BN~\cite{peng2018megdet}). In our Momentum$^2$ Teacher, the student use small-batch statistics to keep efficiency, and the teacher is equipped with a simple and efficient ``Momentum BN" to obtain more stable statistics. }
\label{fig:intro}
\end{figure*}

In the student-teacher framework, as illustrated in Figure~\ref{fig:intro} (a)(b), one image is augmented into two different views for a student and a teacher. The student is trained to predict the representation of the teacher, and the teacher is updated with a ``momentum update" (exponential moving average) of the student. The success of MoCo and BYOL has proven the effectiveness of the student-teacher framework. With the momentum update, the teacher obtains more stable parameters in a temporal ensembling manner~\cite{meanteacher}. 

Besides network parameters, stable statistics~(i.e. batch normalization) is another critical factor for training modern deep networks, especially for self-supervised learning. MoCo~\cite{mocov1} adopts a ``shuffling batch normalization (shuffling BN)'' on the teacher, which shuffles the sample order in the current batch before distributing it among multiple GPUs and rolls back after encoding. This operation prevents the information leaking of the intra-batch communication and still allows training to benefit from BN~\cite{mocov1}. However, shuffling BN only operates on the samples independently for each GPU with small sizes (typical 32 images per GPU), which limits the benefit from larger batch normalization with more stable statistics. 

By leveraging specially designed TPU, BYOL~\cite{BYOL} uses larger batch (\eg, 4096) on both student and teacher to attain more stable statistics, which is equivalent to perform a ``synchronized batch normalization (synced BN)'' on multiple GPUs. The leading performances of BYOL indicates that stable statistics is important. When training with small batch size (\eg, 128 for BN), its performance significantly downgrades~\cite{BYOL}. 
The demand of large batching is unfriendly in many scenarios. Accessing TPU is not common to the research community, and synced BN is very inefficient across multiple GPUs on multiple machines.


In this paper, we propose \emph{Momentum$^2$ Teacher}, which applies momentum update on both network parameters and BN statistics to obtain a more stable teacher. As shown in Figure~\ref{fig:intro} (c), we replace all the batch normalization layers of the teacher with a simple and efficient operation, ``Momentum BN'', to obtain more stable statistics. Momentum BN conducts momentum update on the historical batch statistics~(\emph{i.e.}, mean and standard deviation) to normalize features, instead of just using current mini-batch calculations. This operation enables small batch size training (\eg 128) to generate more stable statistics while keeping the efficiency. Furthermore, as the teacher does not need back-propagation, Momentum BN does not have a gradient issue of the indifferentiable history statistics.
Our method can achieve the state-of-the-art result, without resorting to large batch training on TPU or slow synced BN operation. 

The main contributions of this work are:
\begin{enumerate}[1]
\item We propose a novel approach, named \emph{Momentum$^2$ Teacher}, which keeps efficiency and hardware-friendly of small batch training while achieves competitive performance against large batch training.  
\item The core of our method, Momentum BN, which conducts momentum update on the batch statistics, can benefit all student-teacher based self-supervised methods. It can improves both MoCo and BYOL.
\item We obtain the state-of-the-art result, 74.5\% Top-1 accuracy, on ImageNet under the linear evaluation protocol. 
\end{enumerate}

\section{Related Works}

\paragraph{Self-supervised learning:} Self-supervised approaches have largely reduced the performance gap between supervised models and even achieved superior results on down-stream vision tasks. Contrastive learning measures the (dis)similarities of the sample pairs in a latent space, such as ~\cite{mocov1,mocov2,simclr,infomin,cpc,cpc-1,bachman2019learning,hjelm2018learning,cmc,misra2020self}. Pretext tasks are also heavily researched topic. Some mainly generate pseudo labels by, e.g., clustering features~\cite{caron2018deep,caron2019unsupervised,caron2020unsupervised}, augmenting different views of the single image~(``exemplar'')~\cite{dosovitskiy2014discriminative}, relative and ordered patches~\cite{noroozi2016unsupervised,doersch2015unsupervised,doersch2017multi}, or consecutiveness in videos~\cite{wang2015unsupervised,pathak2017learning}. Others propose to recover the input from corruption, e.g., inpainting~\cite{pathak2016context}, denoising~\cite{vincent2008extracting} and colorization~\cite{zhang2016colorful,zhang2017split} with auto-encoders and GAN. 

\paragraph{Student-teacher framework:}
Mean-teacher~\cite{meanteacher} introduces a student and teacher~(moving-averaged version of the student) network to learn with each other. MoCo~\cite{mocov1,mocov2} combines the contrastive mechanism and the student-teacher framework~\cite{meanteacher}, with a memory bank to maintain a large number of negative samples. BYOL~\cite{BYOL} further removes the negative samples, using an additional predictor to avoid collapsing. It relies on the large number of positive samples, typically 4096 for achieving the best performance. 

\paragraph{Normalization:}
BN~\cite{BN} has widely proven effectively and efficiently in most of the vision tasks. It normalizes the internal feature maps using channel-wise statistics along batch dimension. In practice, BN relies on sufficient batch-size, which is not easily satisfied, especially when large-batch training is acquired in self-supervised learning. There are many techniques proposed to maintain or approximate large batch statistics. Synced BN~\cite{peng2018megdet} increases the batch-size by computing the mean and variance across multiple devices~(GPUs), however, introducing a lot of overhead. Batch Renormalization~\cite{brn} and EvalNorm~\cite{evalnorm} correct batch statistics during training and inference, while compared to Synced BN they generally perform worse. 

Moving averaged batch normalization~\cite{yan2020towards} and online normalization~\cite{online-norm} adopt similar momentum updating of statistics like our momentum BN during the forward pass. However, they need further correct backpropagation for valid SGD optimization, which requires additional computation and memory resources. As there is no back pass within the teacher network, our momentum BN do not need this gradient revision, making it more efficient.     


Another family of normalization is functionally based. Instead of normalizing across samples, layer-norm~\cite{layer-norm} performs normalizes across features which makes it irrelevant to batch-size. Group normalization~\cite{group-norm} and instance normalization~\cite{instance-norm} further extend this by partitioning features into groups. Normalization which is applied to network weights are also proposed, such as weight normalization~\cite{weight-norm} and normalization propagation~\cite{arpit2016normalization}. The very recent study~\cite{byolv2} shows that BYOL even works using functional-based normalization. But the result is still slightly worse than its counterpart like Synced BN~\cite{peng2018megdet}, which is consistent with the observation under supervised training.

\section{Momentum$^2$ Teacher}
We first give our motivation by studying the impact of BN statistics in a series of controlled experiments. Then, we present our method to increase the stability of BN statistics. 

\subsection{Importance of Stable Statistics}
To analysis the role of BN statistics under the student-teacher framework, we design a set of exploratory experiments on STL10~\cite{stl} dataset, based on BYOL baseline~~\cite{BYOL}.

\paragraph{Experiment setup:} 
STL10 contains 96$\times$96 pixel RGB images belonging to ten classes. For each class, it provides 500 images for training and 800 images for testing. We use ResNet18 as the basic network for fast ablation. More training details including image augmentation can be referred in Sec.\ref{sec:iml_detail}.

\begin{table}[h]
\centering
\renewcommand\tabcolsep{2.7pt}
\caption{Comparison on STL10 with different BN statistics in student and teacher. BN and Synced BN indicate the BN statistics are calculated within a single GPU and cross GPUs. All experiments are conducted with 400 epochs. We also report the training speed~(seconds/iteration).}
\begin{tabular}{ccccc}
  \midrule
Student & Teacher & Top1 & Sec./Iter \\
  \midrule
Synced BN         & Synced BN  & 88.06 &  1.1s \\
BN      & Synced BN     & 87.80  &  0.39s \\
Synced BN       & BN       & 87.12 &  0.49s \\
BN & BN       & 84.16 &  0.24s \\
 \midrule
BN      & Momentum BN   & 88.18  & 0.25s \\ 
Synced BN       & Momentum BN   & 88.25  & 0.50s \\
  \midrule
\end{tabular}
\label{tab:impact_of_syncbn}
\end{table}

\paragraph{Observations.} BYOL baseline applies ``Synced BN" (performing BN cross all GPUs) on both student and teacher. Since the teacher does not backprogpagate the gradient, the role of BN in the teacher more replies on the statistics collected in the forward pass. To exam if Synced BN matters, we replace Synced BN with ``BN" (performing BN on single GPU, independently) in student or teacher or both. As showed in Table~\ref{tab:impact_of_syncbn}, we have four observations.

\begin{enumerate}
  \item Synced BN is critical. Without Synced BN, the accuracy significantly drops from 88.06 to 84.16. Similar result was also reported in~\cite{BYOL} when using small batch size. This verified the importance of stable statistics in BN.   
  \item Synced BN slows down the training speed. Because Sycned BN performs cross GPU oprations, the communication overhead is very significant. For example, the speed of Synced BN/Synced BN combination is more than four times slower than BN/BN.
  \item It is not essential to apply Synced BN on both student and teacher. We can get decent results (87.80 or 87.12) with a stable teacher or a stable student. This will enable us to decouple design of BN in student and teacher.
  \item A stable teacher is essential. At the third row in the table, although we did not directly apply Synced BN in the teacher, the teacher still benefit from stable statistics by copying BN parameters in a moving averaged way. Directly applying Synced BN on the teacher is better than on the student (87.80 v.s. 87.12).
\end{enumerate}

Based on the above observations, we conclude that the stable BN statistics brought by large batch training is crucial for student-teacher self-supervised learning. Although stable statistics can be obtained from large-batched samples, it hurts efficiency. Next, we will introduce a new, simple method for higher accuracy and better efficiency.

\subsection{Momentum BN}
Synced BN simply uses large batch to obtain stable statistics. We note that in the student-teacher framework, we do not propagate gradient in the teacher. If we can leverage this characteristic, we may obtain stable statistics using small batch.


We follow the annotation of the batch-normalization. Considering a mini-batch $\mathcal{B}$ of size $m$, then we have $m$ values $x_{1..m}$ of feature. BN operation first calculate two important statistics mean $\mu$ and variance $\sigma$:
\begin{equation}
\begin{aligned}
    \mu &= \frac{1}{m} \Sigma_{i=1}^{m} x_i, \\
    \sigma &= \frac{1}{m} \Sigma_{i=1}^m (x_i - \mu)^2.
\end{aligned}
\end{equation}
Then, BN performs a linear transformation to get final output $y_{1..m}$:
\begin{equation}
\begin{aligned}
    \widehat{x_i} = \frac{x_i - {\mu}}{\sqrt{\sigma + \epsilon}}, \\
    y_i  = \gamma \widehat{x}_i + \beta,
\end{aligned}
\end{equation}
where $\gamma$ and $\beta$ are two learnable parameters.

In the student-teacher framework, two statistics mean $\mu$ and variance $\sigma$ are calculated from samples (encoded by the teacher) in current batch, while $\gamma$ and $\beta$ are momentum version of the student.

Since we do not have to estimate $\gamma$ and $\beta$, and motivated by that the teacher is a kind of temporal ensemble of the student, we perform a momentum update of the BN statistics:
\begin{equation}
\begin{aligned}
    \mu_t &= (1-\alpha) \mu_t + \alpha \mu_{t-1},\\
    \sigma_t &= (1-\alpha) \sigma_t + \alpha \sigma_{t-1},
\end{aligned}
\end{equation}
where $\alpha$ is a momentum coefficient. We simply use an exponential moving average of historical BN statistics to make the teacher more stable. We call the BN with the above momentum update ``\emph{Momentum BN}".

Using momentum statistics is not new in the deep learning. For example, running-mean and running variance have been provided in the deep learning framework like Tensor-Flow for better validation and inference after the training. Here momentum statistics are used for training in the student-teacher framework.


\paragraph{Lazy update.}
Information leaking is one of the main issues in self-supervised learning. For two samples $v, v'$ of the same image, BYOL \emph{sequentially} calculates symmetrized losses as follows:
\begin{equation}
\begin{aligned}
\mathcal{L}_1 &= || \text{student}(v), \text{teacher}(v')||, \\
\mathcal{L}_2 & = || \text{student}(v'), \text{teacher}(v)||.
\end{aligned}
\end{equation}
After calculating the loss $\mathcal{L}_1$, the statistics of $v'$ will be fed into the teacher model. When we calculate the loss $\mathcal{L}_2$, if we use Momentum BN straightforward, we will include the statistics of $v'$ (as historical statistics) into the teacher. This will make the learning more trivial and hurt the performance.

We address this issue by lazily updating Momentum BN statistics. We first perform Momentum BN for L1 and L2 independently, using the statistics of previous batch $\{\mu_{t-1}, \sigma_{t-1}\}$:
\begin{equation}
\begin{aligned}
\{\mu_t^1,\sigma_t^1\} &= (1-\alpha) \{\mu_t^{v'},\sigma_t^{v'}\} + \alpha \{\mu_{t-1},\sigma_{t-1}\},\\
\{\mu_t^2,\sigma_t^2\} &= (1-\alpha) \{\mu_t^v,\sigma_t^v\} + \alpha \{\mu_{t-1},\sigma_{t-1}\},
\end{aligned}
\end{equation}
Then, we conduct BN transformation via $\{\mu_t^1,\sigma_t^1\}$ and $\{\mu_t^2,\sigma_t^2\}$. Last, we update BN statistics:
\begin{equation}
\begin{aligned}
\{\mu_t,\sigma_t\} &= (1-\alpha) \{\frac{\mu_t^{v'}+\mu_t^v}{2},\frac{\sigma_t^{v'},\sigma_t^v}{2}\} + \alpha \{\mu_{t-1},\sigma_{t-1}\}
\end{aligned}
\end{equation}

\paragraph{Results.} We replace Synced BN with Momentum BN in the teacher. As shown in Table~\ref{tab:impact_of_syncbn}, Momentum BN achieves better performance~(88.18 vs 87.8) without requiring cross machine communication. It is as fast as we use small batch BN.

We also apply Momentum BN on the student. The improvement is marginal, verifying that the stable statistics in teacher is essential.

\subsection{Momentum$^2$ Teacher}  
From Table~\ref{tab:impact_of_syncbn}, we can see that student with a small batch (32) statistics already performs closely to that using large-batch (2048) Synced BN, which demonstrates that student can be designed with much smaller batch-size for BN statistics compared to teacher. Therefore, in this paper, we recommend the combination of ``\emph{student with small batch} + \emph{teacher with (small batch) momentum BN}" as our main method. This method has the best trade-off between accuracy and efficiency.

Because the teacher in our method uses momentum mechanism twice, one for weights update from student, the other for calculating BN statistic, we call our method ``\emph{Momentum$^2$ Teacher}".

\subsection{Implementation Details} \label{sec:iml_detail}
\begin{figure}[t!]
\centering
\includegraphics[width=1.0\linewidth, height=0.75\linewidth]{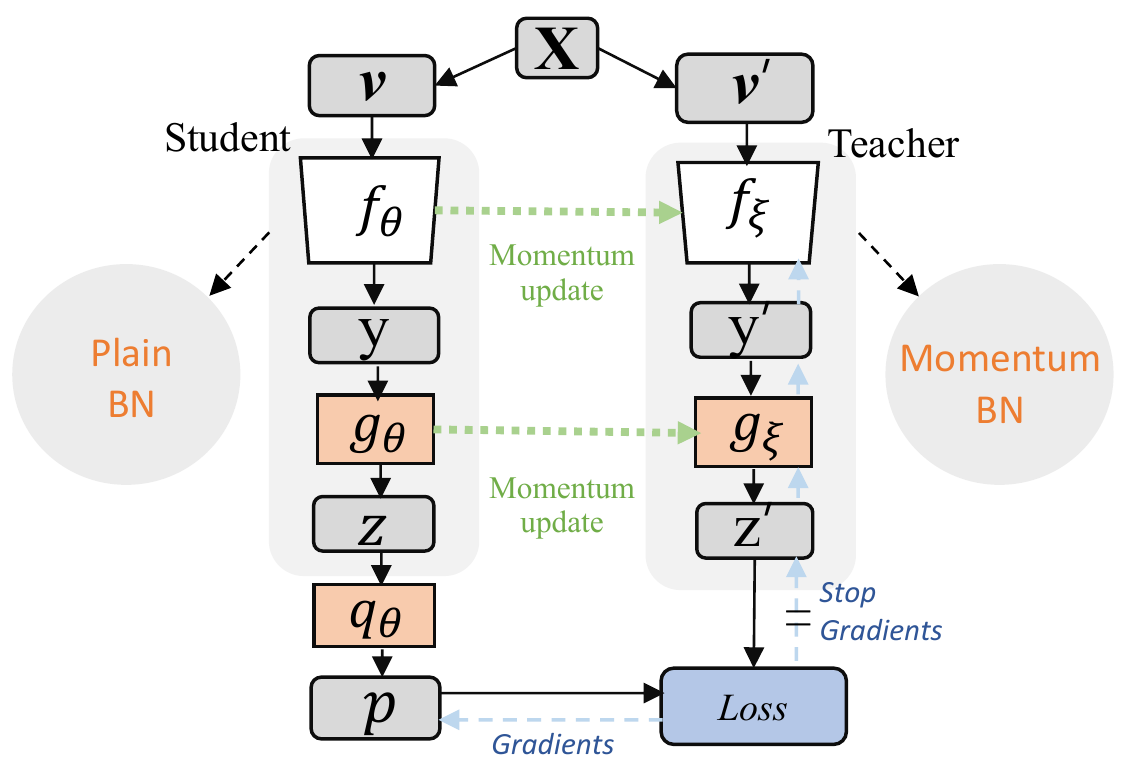} 
\caption{Momentum$^2$ Teacher based on BYOL. Teacher maintains stable BN statistics to improve the performance, and student uses plain BN to keep the efficiency.}
\label{fig:byol}
\end{figure}

\emph{Baseline:} Figure~\ref{fig:byol} shows our method, which uses BYOL~\cite{BYOL} as the baseline. BYOL learns representations by maximizing the similarity between two different augmented views $v$ and $v'$ from the same image $X$. $v$ is passed into the student which consists of a basic encoder $f_\theta$, a MLP $g_\theta$, and a predictor $q_\theta$; and $v'$ is fed into the teacher has only a basic encoder $f_\xi$ and a MLP $g_\xi$. The parameters of $f_\xi$ and $g_\xi$ are momentum update of $f_\theta$ and $g_\theta$:
\begin{equation}
\begin{aligned}
    f_\xi \leftarrow (1 - m) f_\xi + f_\theta,\\
    g_\xi \leftarrow (1 - m) g_\xi + g_\theta.
\end{aligned}
\end{equation}
where $m$ is a momentum coefficient.

\emph{Image augmentations:} We use the same set of image augmentations as in BYOL and SimCLR~\cite{simclr}. Specifically, a crop of fixed size is taken from a random resized and horizontal flipped image, followed by a commonly used color distortion. Then, Gaussian blur and an optional gray-scale are applied to these patches. Finally, solarization is adopted whose probability is set to 0 for student and 0.2 for teacher.

\emph{Architecture:} On STL10, we use ResNet-18~\cite{resnet} as the basic encoder~($f_\theta, f_\xi$), which produces a feature with 512 dimension~($y, y'$) by average pooling. The dimension is set to 512 for first linear layer in the MLP and 128 for the second linear layer. 

\emph{Training:} We train STL10 with 64 2080TI GPUs in order to simulate the cross machine (8 machines) communication in Synced BN. SGD with momentum of 0.9 is adopted without LARS~\cite{lars}. All experiments involve 32 image crops for each GPU unless special statements. At the end of the training, we dump the model from teacher. Learning rate is decay by cosine strategy~\cite{sgdr}. Basic learning rate is set to 0.1. It warm-ups 10 epochs with 0.001 factor and is scaled linearly~\cite{imagenet_in_1_hour} with batch-size. Weight-decay is set to 1e-4. The momentum coefficient of $m$ starts from $m_{base} = 0.032$ and is decreased to zero at the end of the training. In our Momentum BN, the momentum coefficient $\alpha$ starts from $\alpha_{base}=1$ and is decreased to 0 with cosine schedule:
\begin{align}
\label{eqn:momentum}
\begin{split}
m \leftarrow m_{base} \times (cos(\pi k/K) + 1) / 2,\\
\alpha \leftarrow \alpha_{base} \times (cos(\pi k/K) + 1) / 2,
\end{split}
\end{align}
where, $k$ is the current iteration and $K$ is the total number of iterations.

\emph{Evaluation:} We follow the linear classification protocol. The features are frozen and attached by a new linear classifier normalized by BN to fine-tune with given classes.  On STL10, we train 80 epochs using learning rate starting from 0.5~(for 256 batch-size) and decayed with cosine schedule. We train on 8 2080 GPUs using SGD with a momentum of 0.9, without weight-decay.



\section{Experiments}
In this section, we perform training on ImageNet and evaluate the model with the linear evaluation of classification task and some downstream tasks on COCO. Our setup is briefed as follows:

ImageNet~\cite{ImageNet} ILSVRC-2012 dataset has about 1.28 million images belonging to 1000 different classes. The class labels are ignored in the self-supervised learning. The image augmentation is same as STL10's except that image crop uses 224$\times$224.

We use ResNet-50~\cite{resnet} as the encoder. After average pooling, the feature ($y, y'$) has 2048 dimensions. MLP uses a 4098 layer and a 256 layer to generate output ($z, z'$). 

We use 128 2080 GPUs with LARS~\cite{lars} optimizer, but bias and BN weight are excluded from the LARS adaptation and weight-decay. Basic learning rate is set to 0.3, and weight-decay is set to 1.5e-6. The momentum coefficient of parameter $m$ starts from $m_{base} = 0.01$ and is gradually decreased to zero. Fine-tuning uses learning rate 0.2.

\subsection{Effectiveness of Momentum BN}
We first validate the effectiveness of Momentum BN on both BYOL and MoCo frameworks. Table~\ref{tab:byol_with_momentumbn} shows BYOL without Synced BN decreases from 72.5 to 61.5, when we use batch size 128. Our Momentum$^2$ Teacher significantly boosts the performances to 72.9. It is worth noting that Momentum$^2$ Teacher trains as fast as BYOL w/o Synced BN, demonstrating the efficiency (0.6s v.s. 5.25s). 

\begin{table}[h]
\centering
 \caption{Comparison of accuracies between BYOL~\cite{mocov2} and Momentum$^{2}$ Teacher. Following BYOL, we train with 300 epochs. ``BN'' indicates the number of samples to calculate BN.}
\begin{tabular}{l|ccccc}
Model              &  Top1 & Top5 & BN  & Sec./Iter \\
\toprule
BYOL w/ Synced BN  &  72.5 & 87.6 & 4096 & 5.25s \\
BYOL w/o Synced BN &  61.5 & 84.6 & 32 & 0.6s    \\
Momentum$^2$ Teacher   &  72.0 & 90.6 & 32 & 0.6s    \\
Momentum$^2$ Teacher   &  72.9 & 90.6 & 128 & -  \\
\end{tabular}
\label{tab:byol_with_momentumbn}
\end{table}

Then, we replace Shuffling BN in MoCoV2~\cite{mocov2} with our Momentum BN. Following the practice of MoCoV2, we fix the momentum coefficient $\alpha$ in Momentum BN to 0.064. For fast experiment, we train on 32 2080TI GPUs\footnote{Training on 4 machines with 32 GPUs, slightly reduces the MoCo baseline~(trained on 8 GPUs) from 67.5 to 66.8.}. Table~\ref{tab:moco_with_momentumbn} shows that more stable statistics also benefits MoCo, which validates the generality of Momentum BN in the student-teacher framework. Moreover, Momentum BN is near twice faster than shuffling BN. 

\begin{table}[h]
\centering
 \caption{Accuracies of MoCoV2~\cite{mocov2} with Shuffling BN and Momentum BN on ImageNet. Following MoCo, we train with 200 epochs.}
\begin{tabular}{l|ccccc}
   & Top1 & Top5 & Sec./Iter \\
  \midrule
  MoCo w/ Shuffling BN &  66.8 & 87.6 & 0.65s \\
  MoCo w/ Momentum BN &  67.8 & 88.0 & 0.35s \\
\end{tabular}
\label{tab:moco_with_momentumbn}
\end{table}

\subsection{Small Batch-Size}

Next, we compare BYOL with Momentum$^2$ Teacher at different batch sizes and the same training schedule of 300 epochs. We keep using 128 GPUs and changes the number of samples within each GPU. Thus the total batch-sizes will be changed. To achieve stable performances, we extend the linear scaling rule~\cite{imagenet_in_1_hour} by introducing a new equivalence rule for the parameter's momentum coefficient: When the mini-batch size is multiplied by $k$, multiply the learning rates by $k$ and multiply the basic momentum coefficient~$m_{base}$ of parameters by $k$ simultaneously. 

\begin{figure}[!tp]
\centering
\includegraphics[width=1.0\linewidth]{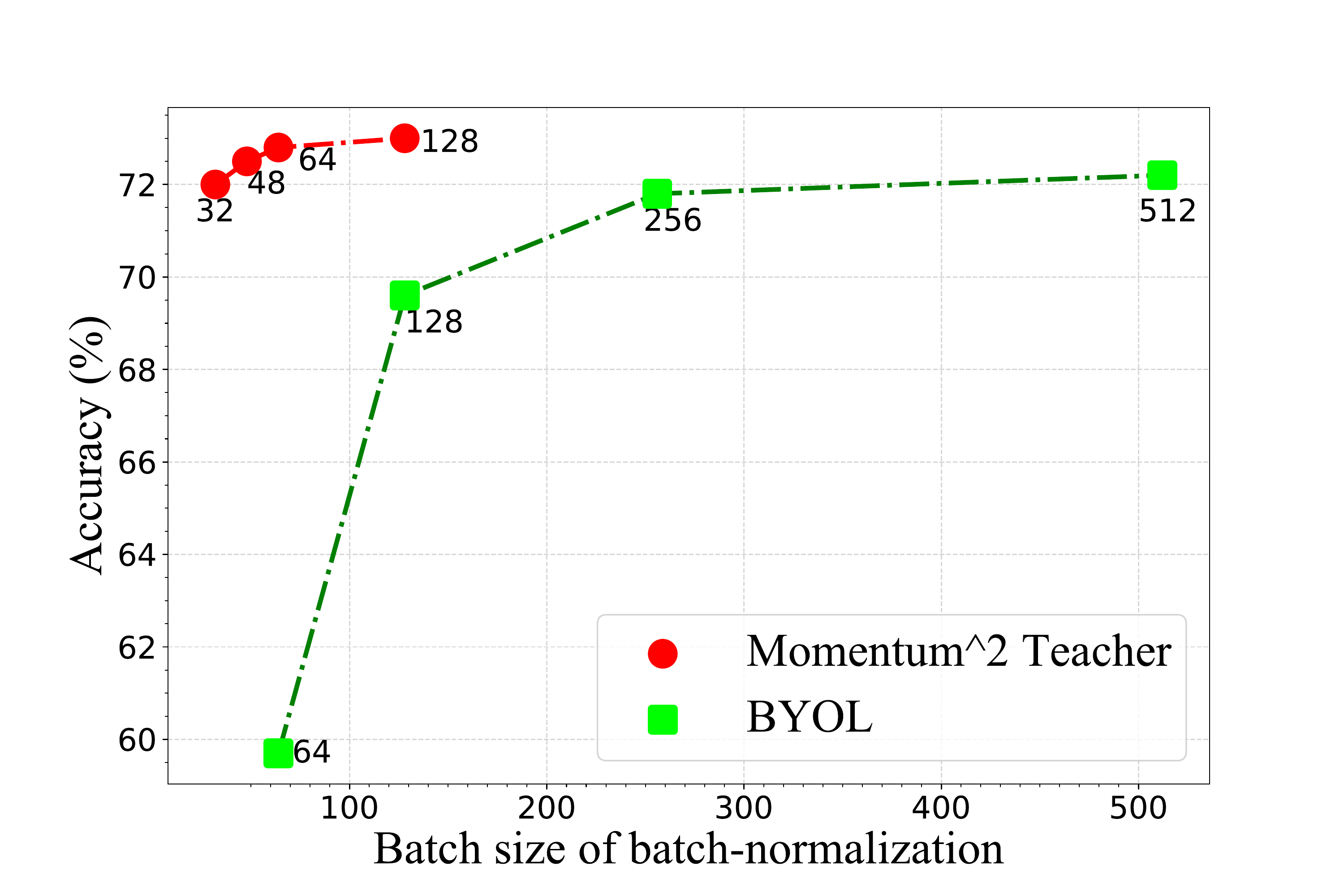} 
\caption{Comparison of using different batch-size to calculate batch-norm in student-network. We plot the `number'' of batch-size to calculate the batch-normalization. BYOL results are referred from its paper. Momentum$^2$ teacher significantly outperforms BYOL when the batch size is reduced. All results are pre-trained with 300 epochs.}
\label{fig:batch_acc}
\end{figure}

\begin{table}[th]
\centering
 \caption{Comparison among different BN batch-sizes~(i.e. the number of samples per GPU) on ImageNet with 300 epochs. BYOL results are referred from its paper conducted by accumulating \emph{N}-steps gradients in parallel. }
\begin{tabular}{l|ccccccc}
  Batch\\size & 16 & 32 & 48 & 64 & 128 & 256 & 512  \\
  \midrule
  BYOL & - & - & - & 59.7 & 69.6 & 71.8 & 72.2 \\
  Ours & 68.3 & 72.0 & 72.5 & 72.6 & 72.9 & - & - \\
  
\end{tabular}
\label{tab:impact_of_bn}
\end{table}

From Table~\ref{tab:impact_of_bn} and Fig.~\ref{fig:batch_acc}, we can see that the number of samples to calculate BN statistics in the student can be very small in our Momentum$^2$ Teacher. Using a batch-size of 32 can still get a decent result, which is on par with BYOL at 512 batch-size. On the contrary, the performance of BYOL rapidly deteriorates when reducing batch size. 

\subsection{Training on a Single Machine}
Thanks to small batch-size training, Momentum$^2$ teacher is also validated on a single machine with 8 GPUs.  We use a learning rate of $lr \times $BatchSize/256~(linear scaling rule~\cite{imagenet_in_1_hour}) with a basic $lr = 0.05$, and then adopt a cosine decay strategy. The weight decay is 0.0001 and the momentum of SGD is 0.9. Large-batch optimizers such as LARS are not involved.

Results are shown in Tab.~\ref{tab:training_on_1_machine}. Compared with large batch-size training, our method achieve even better performances on a single machine with small batch-sizes. We also compare Momentum$^2$ teacher with recent work SimSiam~\cite{simsiam}. Momentum$^2$ teacher consistently outperforms other counterparts at different settings. It is worth noting that our superior results is attained without synced BN operation, which makes it more efficient for training.

\begin{table}[h]
\centering
 \caption{Top 1 accuracies of small batch training on ImageNet linear classification. Our result is obtained on a single machine with 8 V100 GPUs, using batch size of 1024.}
\begin{tabular}{lc|ccc}
Method  &  \begin{tabular}[c]{@{}c@{}}batch\\ size\end{tabular}     & 100e  & 200e  & 300e  \\
\midrule
Ours    &   256       & 70.4 &  - & - \\
Ours    &   1024      & 70.7 & 72.7 &  73.8   \\
SimSiam~\cite{simsiam} &   256     &  68.1 & 70.0 &  -    \\
BYOL~\cite{BYOL}  &   4096     &  -   &  -   &  72.5  \\
\end{tabular}
\label{tab:training_on_1_machine}
\end{table}

\subsection{Parameter Sensitivity}

Momentum$^2$ teacher introduces an additional coefficient $\alpha$ for updating Momentum BN in the teacher. In the previous experiments, $\alpha$ was \emph{dynamically} set from 1 to 0: at the beginning of the training, BN mainly relies on the statistics of the current mini-batch for normalizing; as the training stabilizes, it will rely more on the statistics attained by momentum of historical iterations.

Table~\ref{tab:abl_alpha} compares dynamically adjusted $\alpha$ and fixed $\alpha$. All of fixed $\alpha$ get inferior results. In dynamic adjustment, $\alpha$ will eventually drop to 0, which means it can calculate statistics over the entire data set, making more robust results. Table~\ref{tab:abl_alpha_base} comares the choices of different $\alpha_{base}$~(in Eqn.~\ref{eqn:momentum}). Large $\alpha_{base}$ consistently improves the performance, further validating the early stage of the training should rely more on current batch samples.

\begin{table}[h]
\centering
 \caption{Varying $\alpha$ on ImageNet. All results are pre-trained with 300 epochs.}
\begin{tabular}{l|cccccc}
  $\alpha$ & 1 $\to$ 0 & 1 & 0.5 & 0.2 & 0.1 & 0.01    \\
\midrule
  Top1 & 72.0 & 61.6 & 71.2 &  69.1 & 69.8  & 68.9\\
  Top5 & 90.6 & 84.6 & 90.0 &  89.1 & 89.2  & 88.4 
\end{tabular}
\label{tab:abl_alpha}
\end{table}

\begin{table}[h]
\centering
 \caption{Varying $\alpha_{base}$ on ImageNet. All results are pre-trained with 300 epochs.}
\begin{tabular}{l|cccccc}
  $\alpha_{base}$ & 1 & 0.75 & 0.5 & 0.2 & 0.1 & 0.05 \\
\midrule
  Top1 & 72.0 & 71.6 & 71.2 & 69.7 &  69.5 & 69.6\\
  Top5 & 90.6 & 90.4 & 90.1 & 89.0 &  89.0 & 88.8 
\end{tabular}
\label{tab:abl_alpha_base}
\end{table}

\begin{table}[h]
\centering
 \caption{Comparisons under the linear classification protocol on ImageNet. All methods are instantiated with ResNet-50 basic extractor. We report top-1 and top-5 accuracies in \% on the test set. 
 }
\begin{tabular}{l|cccccc}
  Method & Epoch & Top-1 & Top-5  \\
  \midrule
  Jigsaw~\cite{jigsaw}        & 90     & 45.7 & - \\
  InstDis~\cite{ins_dis}      & 200    & 56.5 & - \\
  BigBiGAN ~\cite{bigbigan}   & -      & 56.6 & - \\
  Local Agg.~\cite{local_agg} & -      & 60.2 & - \\
  CPC v2~\cite{cpc-1}         & 200    & 63.8 & 85.3 \\
  CMC~\cite{cmc}              & -      & 66.2 & 87.0 \\
  SimCLR~\cite{simclr}        & 200    & 66.6 & - \\
  MOCOv2~\cite{mocov2}        & 200    & 67.5 & - \\
  PCL-v2~\cite{pcl}           & 200    & 67.6 & - \\
  InfoMin Aug~\cite{infomin}  & 200    & 70.1 & 89.4 \\
  BYOL~\cite{BYOL}            & 300    & 72.5 & 90.8 \\
  Ours            & 300    & 72.9 & 91.2   \\
  \midrule
  PIRL~\cite{misra2020self}   & 800    & 63.6 & - \\
  SimCLR~\cite{simclr}        & 1000   & 69.3 & 89.0 \\
  MOCOv2~\cite{mocov2}        & 800    & 71.1 & - \\
  InfoMin Aug~\cite{infomin}  & 800    & 73.0 & 91.1 \\
  BYOL~\cite{byolv2}          & 1000   & 74.3 & 91.6 \\
  Ours            & 1000   & 74.5 & 91.7   \\
\end{tabular}
\label{tab:imagenet_linear}
\end{table}

\begin{table*}[h]
     \caption{Fine-tuning of proposal based object detection and instance segmentation on COCO-train2017. We validate on FPN and Cascade R-CNN extended with mask branch. We use ResNet-50-FPN extractor, and report the bounding box AP and mask AP on val2017. 200e and 300e indicate the pre-training epochs.}
     \label{tab:coco_finetuning}
    \begin{subtable}[h]{0.485\textwidth}
        \centering
        \begin{tabular}{l|ccc|ccc}
        pre-train & AP$^{bb}$ & AP$^{bb}_{50} $ & AP$^{bb}_{75}$ & AP$^{mk}$ & AP$^{mk}_{50}$ & AP$^{mk}_{75}$  \\
        \toprule
        random init. & 31.0 & 49.5 & 33.2 & 28.5 & 46.8 & 30.4 \\
        supervised    & 38.9 & 59.6 & 42.7 & 35.4 & 56.5 & 38.1 \\
        \midrule
        MoCo~200e    & 38.5 & 58.9 & 42.0 & 35.1 & 55.9 & 37.7 \\
        BYOL~300e    & 39.6 & 60.9 & 43.3 & 36.7 & 58.0 & 39.3 \\
        \midrule
        Ours~300e   & 39.7  & 61.2  & 43.2   & 36.8   & 58.0 & 39.6    \\
      \end{tabular}
      \caption{Mask R-CNN, \textbf{1}$\times$ schedule}
      \label{tab:coco_fpn_1x}
    \end{subtable}
    \hfill \quad \quad
    \begin{subtable}[h]{0.485\textwidth}
        \centering
        \begin{tabular}{ccc|ccc}
         AP$^{bb}$ & AP$^{bb}_{50} $ & AP$^{bb}_{75}$ & AP$^{mk}$ & AP$^{mk}_{50}$ & AP$^{mk}_{75}$  \\
        \toprule
         36.7 & 56.7 & 40.0 & 33.7 & 53.8 & 35.9 \\
         40.6 & 61.3 & 44.4 & 36.8 & 58.1 & 39.5 \\
        \midrule
        40.8 & 61.6 & 44.7 & 36.9 & 58.4 & 39.7 \\
        41.6 & 62.9 & 45.8 & 38.2 & 59.9 & 41.1 \\
        \midrule
        41.8  & 62.8  & 45.9   & 38.4   & 60.1  & 41.2  \\
      \end{tabular}
      \caption{Mask R-CNN,  \textbf{2}$\times$ schedule}
      \label{tab:coco_fpn_2x}
     \end{subtable}
    \vfill
    \begin{subtable}[h]{0.485\textwidth}
        \centering
        \begin{tabular}{l|ccc|ccc}
        pre-train & AP$^{bb}$ & AP$^{bb}_{50} $ & AP$^{bb}_{75}$ & AP$^{mk}$ & AP$^{mk}_{50}$ & AP$^{mk}_{75}$  \\
        \toprule
        random init.  & 35.3 & 51.0 & 38.3 & 31.0 & 48.8 & 33.3 \\
        supervised    & 42.1 & 59.8 & 45.9 & 36.4 & 57.1 & 39.3 \\
        \midrule
        MoCo~200e    & 42.4 & 59.8 & 46.1 & 37.0 & 57.2 & 40.1 \\
        BYOL~300e    & 43.6 & 61.5 & 47.5 & 38.3 & 59.0 & 41.6 \\
        \midrule
        Ours~300e   & 43.6  & 61.6  & 47.3   & 38.2   & 58.8 & 41.2    \\
      \end{tabular}
      \caption{Cascade R-CNN,  \textbf{1}$\times$ schedule}
      \label{tab:coco_fpn_1x}
    \end{subtable}
    \hfill \quad \quad
    \begin{subtable}[h]{0.485\textwidth}
        \centering
        \begin{tabular}{ccc|ccc}
         AP$^{bb}$ & AP$^{bb}_{50} $ & AP$^{bb}_{75}$ & AP$^{mk}$ & AP$^{mk}_{50}$ & AP$^{mk}_{75}$  \\
        \toprule
         40.7 & 57.7 & 44.2 & 35.9 & 55.5 & 39.1 \\
         43.5 & 61.3 & 47.4 & 37.7 & 58.7 & 40.8 \\
        \midrule
        44.3 & 61.9 & 48.1 & 38.7 & 59.5 & 42.0 \\
        45.0 & 62.9 & 48.8 & 39.3 & 60.5 & 42.6 \\
        \midrule
        45.1  & 63.0  &49.0   & 39.4   & 60.7  & 42.8  \\
      \end{tabular}
      \caption{Cascade R-CNN, \textbf{2}$\times$ schedule}
      \label{tab:coco_fpn_2x}
     \end{subtable}
\end{table*}

\subsection{Comparison with State-of-the-art}

In this section, we first compare performances of Momentum$^2$ Teacher's representation with recent state-of-the-art self-supervised approaches on ImageNet. As the main merit of self-supervised learning is to learn \emph{transferrable} feature, we then measure the transfer capabilities on COCO~\cite{mscoco} and LVIS~\cite{lvis} dataset which includes both annotations for object detection and segmentation. To achieve better results, the pre-training uses 128 samples within each GPU.


\subsubsection{Linear Evaluation on ImageNet}

As shown in Table ~\ref{tab:imagenet_linear}, our method (ResNet-50 based) obtains 72.9\% and 74.5 top-1 accuracies under 300 and 1000 epochs, which outperform the previous state of the arts. It is worth noting that we only adopt 128 samples for batch-normalization within student, while BYOL requires 4096. Our superior result is also achieved without any additional architecture requirement and stronger augmentation, keeping it simple and effective for practice.

Recent SwAV~\cite{caron2020unsupervised} method can achieve higher accuracy 75.3\% by using additional 6 crops within each training iteration to, at the cost of more forward computations.

\begin{table*}[h]
     \caption{Fine-tuning of dense object detection on COCO-train2017. We validate on RetinaNet and FCOS with ResNet-50-FPN and report the AP of bounding boxes.}
     \label{tab:coco_finetuning_dense_detector}
    \begin{subtable}[h]{0.32\textwidth}
        \centering
        \begin{tabular}{l|ccc}
        pre-train & AP$^{bb}$ & AP$^{bb}_{50} $ & AP$^{bb}_{75}$  \\
        \toprule
        random init.  & 24.4 & 38.8 & 25.8  \\
        supervised    & 37.3 & 56.6 & 39.8  \\
        \midrule
        MoCo~200e    & 37.2 & 56.4 & 40.0  \\
        BYOL~300e    & 36.4 & 55.9 & 39.2 \\
        \midrule
        Ours~300e   & 35.9  & 55.3  & 38.4 \\
      \end{tabular}
      \caption{RetinaNet, \textbf{1}$\times$ schedule}
      \label{tab:coco_fpn_1x}
    \end{subtable}
    \hfill
    \begin{subtable}[h]{0.2\textwidth}
        \centering
        \begin{tabular}{ccc}
         AP$^{bb}$ & AP$^{bb}_{50} $ & AP$^{bb}_{75}$  \\
        \toprule
         31.2 & 48.1 & 33.3  \\
         38.7 & 58.2 & 41.4  \\
        \midrule
        39.0 & 58.3 & 41.6  \\
        39.5 & 59.3 & 42.6  \\
        \midrule
        39.4 & 59.3 & 42.2   \\
      \end{tabular}
      \caption{RetinaNet, \textbf{2}$\times$ schedule}
      \label{tab:coco_fpn_2x}
    \end{subtable}
        \hfill
    \begin{subtable}[h]{0.2\textwidth}
        \centering
        \begin{tabular}{ccc}
         AP$^{bb}$ & AP$^{bb}_{50} $ & AP$^{bb}_{75}$  \\
        \toprule
         25.0 & 39.2 & 26.4  \\
         38.7 & 57.6 & 41.9 \\
        \midrule
        39.0 & 57.5 & 42.0  \\
        37.6 & 56.0 & 41.0   \\
        \midrule
        37.7 & 56.8 & 40.8 \\
      \end{tabular}
      \caption{FCOS, \textbf{1}$\times$ schedule}
      \label{tab:coco_fpn_2x}
    \end{subtable}
        \hfill
    \begin{subtable}[h]{0.2\textwidth}
        \centering
        \begin{tabular}{ccc}
         AP$^{bb}$ & AP$^{bb}_{50} $ & AP$^{bb}_{75}$  \\
        \toprule
         31.8 & 48.2 & 33.6  \\
         38.5 & 57.0 & 41.3   \\
        \midrule
        38.8 & 57.1 & 41.7  \\
        39.5 & 58.3& 42.5    \\
        \midrule
        39.9  & 59.0  & 43.2  \\
      \end{tabular}
      \caption{FCOS, \textbf{2}$\times$ schedule}
      \label{tab:coco_fpn_2x}
    \end{subtable}
\end{table*}

\subsubsection{Transferring Features}

For COCO dataset, we fine-tune on the \emph{train2017}~(about 118k images) set and tested on \emph{val2017}~(5k). The typical FPN~\cite{fpn} backbone is adopted and further cooperated with Synced BN. The short-edge of training image is in [640, 800] sampled with 32 pixels intervals, while fixed at 800 during testing. We follow the typical 1$\times$ or 2$\times$ training strategy provided by Detectron2~\cite{detectron2} repository. 
For LVIS 0.5 dataset, we uses 56K images over 1230 categories for training, and 5k images for validation.

\paragraph{Proposal based Detector and Segmentor}
We adopt FPN and Cascade R-CNN extended with mask branch~\cite{fpn,mask_rcnn,cascade_rcnn} to validate the transfer capabilities on COCO object detection and segmentation. Following common practice, we stack 4 $3 \times 3 $ convolutions in the mask branch and utilize 2 linear layers in the detection branch. 

Table~\ref{tab:coco_finetuning} shows the bounding box AP and mask AP on COCO \emph{val2017}. For both two-stage and multi-stage instance segmentor, our method significantly outperforms the supervised counterpart, achieving comparable results as BYOL with more efficient implementation. Meanwhile, we surpass another student-teacher based self-supervised method, MoCo, which uses a small-batch of 32 by a large margin.

\paragraph{Dense Object Detector}
Next we compare the representation by fine-tuning dense object detectors~(a.k.a single-stage object detector). We adopt two typical dense detector, namely anchor-based RetinaNet~\cite{retinanet} and anchor-free FCOS~\cite{fcos}. We follow the baseline provided by Detectron2. Particularly, FCOS utilizes group normalization~\cite{group-norm} for convolutions in detection head, while RetinaNet does not. Moreover, RetinaNet adopt the standard multi-scale training setting while FCOS trains with a single-scale of 800 pixels.

Table~\ref{tab:coco_finetuning_dense_detector} shows the bounding box AP of the detectors initialized with different pre-training weights. Our method shows great potential when training with a long schedule~(2$\times$). 
We achieve comparable performance as BYOL on RetinaNet while yield much better results on FCOS, outperforming the supervised counterpart. We observe that both BYOL and our method get inferior results when training with the 1$\times$ schedule, which we leave its exploration for future work.

\paragraph{LVIS Instance Segmentation} We further transfer the instance segmentation on LVIS which contains about 1000 long-tailed distributed categories. Our method significantly outperforms the supervised pre-training by an mAP of 1.7\% in Tab.~\ref{tab:coco_lvis}, validating the generalization of our approach. We hypothesis that self-supervised pre-training can benefit more for the task with fewer annotated samples.

\begin{table}[h]
     \caption{Fine-tuning of Mask R-CNN  on LVIS 0.5. All experiments are instantiated with ResNet-50-FPN and trained with 2$\times$ schedule.}
     \label{tab:coco_lvis}

        \centering
        \begin{tabular}{l|ccc}
        pre-train & AP$^{mk}$ & AP$^{mk}_{50} $ & AP$^{mk}_{75}$   \\
        \toprule
        random init.  & 22.5 & 34.8 & 23.8 \\
        supervised    & 24.4 & 37.8 & 25.8 \\
        \midrule
        MoCo~200e    & 24.1 & 37.4 & 25.5 \\
        Ours~300e    & 26.1~(+1.7) & 39.8~(+2.0) & 28.1~(+2.3)  \\
      \end{tabular}

\end{table}

\begin{table}[h]
\centering
 \caption{Accuracy of models trained with few image labels. We report top-1 and top-5 accuracies in \% on the test set. }
\begin{tabular}{lccc|cc}
  \multirow{ 2}{*}{Method} & \multirow{ 2}{*}{epoch} & \multicolumn{2}{c}{Top-1} & \multicolumn{2}{c}{Top-5}  \\
       &  & 1\% & 10\% & 1\% & 10\%  \\
  \toprule
Supervised\cite{zhai2019s4l} & - & 25.4 & 56.4 & 48.4 & 80.4 \\
\midrule
 \emph{Semi-supervised:} \\ 
\midrule
 Pseudolabels~\cite{s4l} & - & - & - & 51.6 & 82.4\\
 VAT~\cite{VAT}        & - & - & - & 47.0 & 83.4 \\
 UDA~\cite{UDA}      & - & - & 68.8 & - & 88.5 \\
 FixMatch~\cite{fixmatch}  & - & - & 71.5 & - & 89.1\\
 \midrule
\emph{Self-supervised:} \\
\midrule
InstDisc~\cite{ins_dis} &  200 & -  & - & 39.2 &  77.4 \\
PIRL~\cite{misra2020self}   & 800 &  - & -  & 57.2 & 83.8\\
MoCov2~\cite{mocov2} & 800 & 42.3 & 63.8 & 70.1 &  86.2\\
PCL~\cite{pcl} & 200 & - &  - & 75.3 & 85.6 \\
SimCLR~\cite{simclr} & 1000 & 48.3 & 65.6 & 75.5 & 87.8  \\
SWAV(B=256)\cite{caron2020unsupervised} & 200 & 51.3 &  67.8 & 76.6 & 88.6 \\ 
SWAV(B=4096)\cite{caron2020unsupervised} & 200 & 52.6 &  68.5 & 77.7 &  89.2\\ 
SWAV(B=4096)\cite{caron2020unsupervised} & 800 & 53.9 &  70.2 &  78.5 &  89.9\\ 
BYOL~\cite{BYOL}  & 1000 &  53.2 & 68.8 & 78.4 & 89.0 \\
Ours & 300 &  57.7 & 70.2 & 80.8 & 89.3\\
Ours & 1000 &  62.3 & 72.2 & 84.1 & 90.1 \\
\end{tabular}
\label{tab:imagenet_linear_semi}
\end{table}

\subsection{Semi-Supervised Training on ImageNet} Last, we evaluate the classification performance obtained when fine-tuning ours representation using a small subset of the ImageNet's train set with label information. We use the same fixed splits of 1\% and 10 \% provided by SimCLR.  The training most follows common semi-supervised protocol in \cite{simclr,BYOL,cpc-1}, we perform training on 8 V100 GPUs with total batch-size of 1024. Learning rate is set to 0.08 and decayed by cosine strategy. We fine-tune 20/15 epochs for splits of 1\% /10\% respectively. After training, the statistics of BN are recounted for the later testing. We report the top-1 and top-5 accuracies on the test-set in Tab.~\ref{tab:imagenet_linear_semi}. Our method consistently outperforms the previous approaches.

\section{Conclusion}
We have presented Momentum$^{2}$ Teacher, for self-supervised learning, by introducing a simple and efficient Momentum BN operation on the teacher. With a more stable teacher, we are able to use fast small-batch training to obtain the leading results, which is more friendly to the majority of the researchers.

{\small
\bibliographystyle{ieee_fullname}
\bibliography{egbib}
}

\end{document}